\documentclass[11pt,a4paper]{article}
\usepackage[hyperref]{emnlp2020}
\usepackage{times,graphicx,subcaption}
\usepackage{latexsym}
\usepackage{enumitem}
\newcommand{\textbi}[1]{\textbf{#1}}

\usepackage{microtype}

\aclfinalcopy 

\newenvironment{itemize*}%
  {\begin{itemize}%
    \setlength{\itemsep}{0.9pt}%
    \setlength{\parskip}{0.9pt}%
    \setlength{\topsep}{0.9pt}}%
  {\end{itemize}}

\title{What Can We Do to Improve Peer Review in NLP?}

\author{Anna Rogers \\
  Center for Social Data Science \\
  University of Copenhagen \\
  \texttt{arogers@sodas.ku.dk} \\\And
  Isabelle Augenstein \\
  Department of Computer Science \\
  University of Copenhagen \\
  \texttt{augenstein@di.ku.dk} \\}

\date{}

\begin{document}
\maketitle
\begin{abstract}
Peer review is our best tool for judging the quality of conference submissions, but it is becoming increasingly spurious. We argue that a part of the problem is that the reviewers and area chairs face a poorly defined task forcing apples-to-oranges comparisons. There are several potential ways forward, but the key difficulty is creating the incentives and mechanisms for their consistent implementation in the NLP community.
\end{abstract}

\section{Introduction}

Traditionally, peer review is expected to act as a filter for high-quality, impactful work \cite{Wingfield_2018_peer_review_system_has_flaws_But_its_still_barrier_to_bad_science}, but this does not hold in practice: 

\begin{itemize}[noitemsep,topsep=0pt]
    \item \textit{Peer review does not guarantee quality control}, neither for small errors nor for serious methodological flaws -- even in biomedical literature, where publishing flawed results does real damage \cite{Smith_2010_Classical_peer_review_empty_gun}. 
    \item \textit{Peer review fails to detect impactful papers}. The correlation between conference rejection rates and conference impact in terms of citations is not strong \cite{FreyneCoyleEtAl_2010_Relative_status_of_journal_and_conference_publications_in_computer_science,RagoneMirylenkaEtAl_2011_quantitative_analysis_of_peer_review,RagoneMirylenkaEtAl_2013_On_peer_review_in_computer_science_Analysis_of_its_effectiveness_and_suggestions_for_improvement}, and rejects from one conference sometimes receive awards\footnote{\citet{Mani_2011_Improving_our_reviewing_processes} discusses the example of a paper by Branavan et al. that received the award at ACL 2009 after being rejected from NAACL (scored at 2.3/5). More recently, ELMo \cite{PetersNeumannEtAl_2018_Deep_Contextualized_Word_Representations} received low scores from ICLR reviewers and was resubmitted to NAACL to win the award there.} at another. 
\end{itemize}

The problem is that both expectations are unrealistic to begin with. A peer reviewer cannot perform real quality control, because that would mean ensuring that a paper is \textit{reproducible}. Not only is that impossible, only having a few hours to review a paper, but it is a general problem for Deep Learning (DL)-based NLP \cite{Crane_2018_Questionable_Answers_in_Question_Answering_Research_Reproducibility_and_Variability_of_Published_Results,Rogers_2019_How_Transformers_broke_NLP_leaderboards}. The reproducibility checklist at EMNLP 2020 \cite{DodgeGururanganEtAl_2019_Show_Your_Work_Improved_Reporting_of_Experimental_Results} is the first step in that direction.

As for paper impact, it is distinct from its scientific merit \cite{BhattacharyaPackalen_2020_Stagnation_and_Scientific_Incentives}, and strongly depends on completely orthogonal factors: how niche is the topic, how much promotion was done, whether the paper offers room to innovate with a low entry barrier\footnote{Amongst the biggest success stories in DL-based NLP are word2vec \cite{MikolovChenEtAl_2013_Efficient_estimation_of_word_representations_in_vector_space} and BERT \cite{DevlinChangEtAl_2019_BERT_Pre-training_of_Deep_Bidirectional_Transformers_for_Language_Understanding} Note that both of them contributed a transfer learning paradigm with room for incremental modifications.} \cite{Anderson_2009_Conference_reviewing_considered_harmful}.

What we \textit{could} realistically expect from peer review is to reject the papers with obvious methodology flaws, and turn the spotlight on the ideas which would be beneficial for the field to discuss. However, the current process is not set up to achieve either purpose. Instead, it aims to rank all submissions by their merit so as to identify the top ~25\%.
That task, we argue, is fundamentally impossible. 

\begin{figure}[!t]
    \centering
    \includegraphics[width=.6\columnwidth]{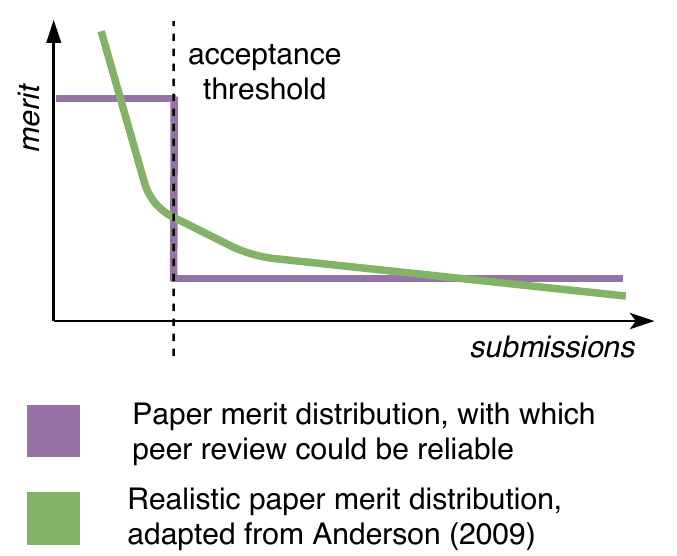}
    \caption{Paper merit distribution}
    \label{fig:merit-distribution}
\end{figure}

\section{Why Is Peer Review So Difficult?}

Peer review would be easy if the paper merit distribution had a clear boundary between good and bad papers (and ideally that boundary would match the conference acceptance threshold). However, that is clearly not the case. Based on citation counts, \citet{Anderson_2009_Conference_reviewing_considered_harmful} hypothesize that paper merit is Zipf-distributed, as shown in \autoref{fig:merit-distribution}. That means that even with the most objective reviewers, the difference between the worst accepted paper and the best rejected paper is less than 1\%.

To make matters worse, there are no clear criteria that would help to draw the decision boundary. \citet{Anderson_2009_Conference_reviewing_considered_harmful} discusses an experiment at SIG-COMM 2006, where they first made the easy accept/reject decisions for papers with low review score variance, and then assigned up to 9 additional reviewers to papers with high variance. The reviewers who had to discuss the difficult cases were reportedly ``nearly driven insane" by the apples-to-oranges comparisons, such as incomplete evaluation in one borderline paper vs narrow applicability of another. No matter how long we agonize over such decisions, they will look random. A case in point: at NIPS 2014, 10\% submissions were reviewed by two different PCs, who disagreed on 57\% of papers \cite{Price_2014_NIPS_experiment}.

\begin{figure}[!t]
    \centering
    \includegraphics[width=.7\linewidth]{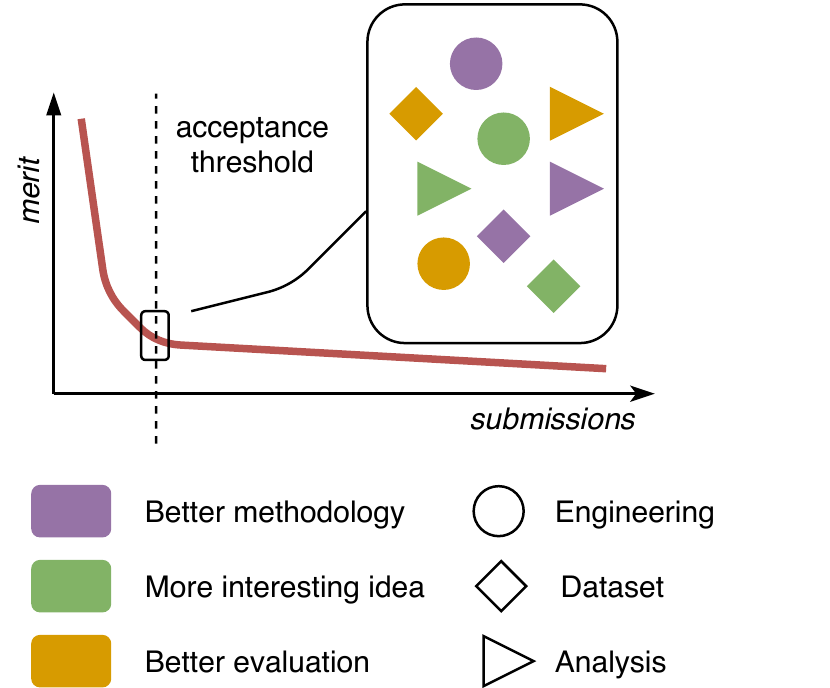}
    \caption{Why it is hard to compare borderline papers}
    \label{fig:criteria}
\end{figure}

For large *ACL conferences, the situation is even worse: we often weigh against each other different \textit{types} of papers with different strengths and weaknesses (\autoref{fig:criteria}). There can be no `correct' answer as to which one has more scientific merit. 

\section{How Reviewers Cope}

Faced with an objectively impossible task, reviewers do what humans generally do to reason under uncertainty: they default to heuristics, which introduces unwanted biases \citep{korteling2018neural}. There is an extra incentive to do so because apples-to-oranges comparisons are a slow, deliberate, cognitively expensive process 
-- and peer review is currently invisible work performed for free. 

This section lists some of the most problematic reviewer heuristics in NLP.  

\textbi{Writing style.} Language errors, non-standard style or rhetorical structure are easy to spot and interpret as sloppiness. This puts almost everybody at a disadvantage against North Americans. 
Papers with worse English may even be perceived as worse than those with better content \cite{Church_2020_Emerging_trends_Reviewing_reviewers_again}. 

\textbi{Results not surpassing SOTA.} 
An easy heuristic is to check if the paper beats the state of the art (SOTA).
While an engineering contribution should demonstrate a significant improvement over prior methods, it does not have to be an improvement in \textit{predictive performance}: advances in compute or data efficiency, interpretability, cognitive plausibility etc. are also valuable \cite{Rogers_2020_Peer_review_in_NLP_reject-if-not-SOTA}.
The focus on predictive performance encourages the `arms race' for pre-training data and compute, and exacerbates methodological issues\footnote{Inter alia: unfair comparisons \cite{MusgraveBelongieEtAl_2020_Metric_Learning_Reality_Check}, dependence on non-architecture-related factors \cite{DodgeIlharcoEtAl_2020_Fine-Tuning_Pretrained_Language_Models_Weight_Initializations_Data_Orders_and_Early_Stopping,Crane_2018_Questionable_Answers_in_Question_Answering_Research_Reproducibility_and_Variability_of_Published_Results}, no incentives for producing robust systems \cite{EthayarajhJurafsky_2020_Utility_is_in_Eye_of_User_Critique_of_NLP_Leaderboards}, flawed benchmarks \cite{JiaLiang_2017_Adversarial_Examples_for_Evaluating_Reading_Comprehension_Systems,McCoyPavlickEtAl_2019_Right_for_Wrong_Reasons_Diagnosing_Syntactic_Heuristics_in_Natural_Language_Inference}, which become a tool for producing incremental papers \cite{Reiter_2020_Why_do_we_still_use_18-year_old_BLEU}}. The requirement for comparisons with the latest SOTA model also puts us in the hamster wheel, making experiments outdated already by the submission time.

\textbi{Narrow topics.} It is easier to publish on trendy, `scientifically sexy' topics \cite{Smith_2010_Classical_peer_review_empty_gun}. In the last two years, there has been little talk of anything other than large pretrained Transformers, with BERT alone becoming the target of over 150 studies proposing analysis and various modifications \cite{RogersKovalevaEtAl_2020_Primer_in_BERTology_What_we_know_about_how_BERT_works}. The `hot trend' forms the prototype for the kind of paper that should be recommended for acceptance. Niche topics such as historical text normalization are downvoted (unless, of course, BERT could somehow be used for that).

\textbi{Work not on English.} Since prototypical NLP experiments use English as the target language, other languages mark the paper as narrow. This heuristic is indefensible, since approaches tested only on e.g. Estonian are as generalizable as those tested only on English. It also strengthens the `default' status of English \cite{Bender_2019_BenderRule_On_Naming_Languages_We_Study_and_Why_It_Matters}.

\textbi{Already-famous work and work from well-known labs.} If reviewers feel that a paper was already accepted by the research community, they do not need to do any more vetting. For example, there was no way for BERT to go through fully anonymous peer review \cite{Cotterell_2019_We_should_Anonymize_Model_Names_during_Peer_Review}. 

Early preprint citations are arguably a better indicator of paper quality than peer review \cite{Church_2017_Emerging_trends_I_did_it_I_did_it_I_did_it_but,Church_2020_Emerging_trends_Reviewing_reviewers_again}, but they are also influenced by how famous the authors are\footnote{\citet{PetersCeci_1982_fate_of_published_articles_submitted_again} resubmitted 12 articles to psychology journals that already \textit{published} these articles, with the author names changed to unknown names. Many were rejected for `methodology flaws'! See \citet{Rogers_2020_anonymity} for discussion of anonymity in upcoming ACL peer review reform.}, and how much they publicize their work. Well-known labs tend to have large online followings or even PR departments, propagating the `rich get richer' phenomenon (also known as the `Matthew Effect', \citet{Merton1968.20140805}). 

\textbi{The proposed solution seems too simple.} Since a prototypical `acceptable' paper features a sophisticated DL model, simple solutions may look like the authors did not do enough work. This is obviously flawed, as the goal is to solve the problem, rather than to solve it in a complex way. 

\textbi{Non-mainstream approaches.} Since a `mainstream' *ACL paper currently uses DL-based methods, anything else might look like it does not really belong in the main track - even though ACL stands for `Association for Computational Linguistics'. That puts interdisciplinary efforts at a disadvantage, and continues the trend for intellectual segregation of NLP \cite{Reiter_2007_shrinking_horizons_of_computational_linguistics}. E.g., theoretical papers and linguistic resources should not be a priori at a disadvantage just because they do not contain DL experiments.

\textbi{Resource papers.} Surprising as it may seem in a field that relies so much on supervised machine learning, resource papers are routinely rejected simply for being resource papers. Linguistically deeper papers may also receive extra penalties for linguistic details at the cost of DL experiments, for non-English resources, and, by analogy with the SOTA heuristic for engineering papers, for not offering the \textit{largest} resource \cite{Rogers_2020_Peer_review_in_NLP_resource_papers}.

\textbi{Novel approaches.} This sounds almost absurd, but scientific peer review is systematically biased towards unobjectionable (rather than novel) work \cite{Church_2005_Reviewing_reviewers,Church_2020_Emerging_trends_Reviewing_reviewers_again,Smith_2010_Classical_peer_review_empty_gun,BhattacharyaPackalen_2020_Stagnation_and_Scientific_Incentives}. A reviewer faced with evaluating a completely new idea without prior art has to make a more difficult call than one for a paper with clear predecessors and a leaderboard table, and is more likely to fall back to one of the heuristics. The very process based on majority votes necessarily promotes `safe', incremental, likely boring work, and  puts non-mainstream work at a disadvantage.

\textbi{Substitute questions.} The question \textit{``how good is this paper?"} is difficult, because the criteria for scientific merit are vague. What humans often do to answer a difficult question is to unconsciously substitute it with an easier one \cite{Kahneman_2013_Thinking_fast_and_slow}. We suspect that one of these substitutes is \textit{``are there any obvious ways to improve this paper?''} This would explain the acceptance rate gap for long and short papers:\footnote{In 2020: 24.6\% long vs 16.7\% short papers at EMNLP, 25.4\% vs 17.6\% at ACL, 35.5\% vs 27.7\% at COLING.} since the latter include fewer details and experiments, they are easier to find fault with. In our experience, another such substitute question may be \textit{``if I did this study, would I make the same choices?"} The reviewers using this heuristic are not actually responding to the real paper they were assigned, but to an imaginary paper more in line with their interests and methodology -- and the real paper compares unfavorably.

EMNLP 2020 explicitly addressed most of the above heuristics in its blog \cite{LiuCohnEtAl_2020_Advice_on_Reviewing_for_EMNLP}, but naming and shaming is unlikely to be sufficient. Heuristics \textit{are} the way humans reason under uncertainty, so the only way to fix this is to clarify the very task reviewers are expected to perform.

\section{Can We Just Abolish Peer Review?}

If the task is fundamentally impossible, should we just give up and look for alternatives to peer review? Each round of conference notifications spurs calls on social media to just abolish the whole system, to increase acceptance rate, to let citations be the metric of the paper quality. 

Unfortunately, this is not realistic, even if there were no co-dependence between citation counts and scientific fame or promotion efforts. Fundamentally, low acceptance rates are a proxy for paper quality for non-experts, and that metric is expected by almost every hiring and grant committee. We are not aware of serious proposals on how to change that. And any experiments will require a generation of extremely brave students who are willing to graduate with no `respectable' publication record.

EMNLP 2020 essentially increased the acceptance rate by creating a second-tier publication named \textit{Findings of EMNLP}, which has ``no requirement for high perceived impact, and accordingly solid work in untrendy areas and other more niche works will be eligible''.\footnote{\url{https://2020.emnlp.org/blog/2020-04-19-findings-of-emnlp/}} It enabled the organizers to accept 15.5\% of extra submissions (including this one), in addition to 22.4\% in the main track. 

Unfortunately, this approach does not address the fundamental issue (comparing the incomparable), and introduces new problems:

\begin{itemize}[noitemsep, topsep=0pt]
    \item the very existence of \textit{Findings} is likely to exacerbate reviewer biases: they may give lower scores to `non-trendy' work to nudge it towards \textit{Findings}, even if not explicitly asked for main track vs \textit{Findings} recommendations);
    \item no matter what status \textit{Findings} attains in the community, in the academic rankings it will always remain a second-tier outlet, and that will change trajectories of careers and grants of people who engage in `non-trendy' work;
    \item \textit{Findings} implicitly caters to `fast science': rather than improving a paper, authors can publish it as is and move on. In the short run, this helps the authors (particularly those whose SOTA results are likely to `expire'). In the long run, it means more papers which are less well executed. 
\end{itemize}

Finally, \textit{Findings} also decreases the likelihood that a new top-tier venue would emerge to make the `untrendy' topics trendy, and potentially change the direction of the field. Ironically, EMNLP itself was born as a home to papers rejected by conservative ACL reviewers \cite{Church_2005_Reviewing_reviewers}. If ACL had created \textit{Findings} in 1996, there would likely be no EMNLP today -- and the whole field might be less empirical.

\section{So What Can We Do?}

Until there are systemic changes in how researchers are evaluated, peer review remains `the least bad system available' \cite{Smith_2010_Classical_peer_review_empty_gun}. Still, there is clearly room for improvement.

First, peer review has to become a valued part of academic CVs, and something that employers budget time for. Reviews done by overworked people in their free time will not be top-quality.

Second, we need to reduce the need for reviewers and ACs to reason under high uncertainty. It cannot be fully eliminated, but there are several obvious directions for improvement. 

    \textbi{Better reviewer matching}. Reviewers are more likely to resort to heuristics when they are not experts in the same narrow area as the paper they are reviewing. A matching should take into account both the tasks and the methods (e.g. a paper on coreference annotation is unlikely to be appreciated by a practitioner who only worked on coreference applications). Since it is not always possible to find perfect matches, reviewers with complementary partial expertise (e.g. someone who speaks the language if the paper is not on English, plus an area expert) could be a fall-back strategy.
    
    \textbi{More fine-grained tracks}: ACs should never have to decide between different types of papers. If surveys, opinion pieces, resource and analysis papers etc. are all welcome, they should have their own acceptance quotas and best paper awards.
    
    \textbi{Review forms tailored for different paper types:} it does not make much sense to evaluate a reproducibility report for novelty, or a resource paper for SOTA results. COLING 2018 developed review forms taking into account different types of contributions, possibly several per paper.

\textbi{Announcing editorial priorities pre-submission}. What is the primary focus of a particular conference: SOTA engineering, diversity of approaches, fresh ideas? What counts more towards acceptance? Stating this clearly would help authors find an appropriate event for their work, and help reviewers and area chairs be more consistent in their recommendations.

\textbi{Not asking the reviewers for overall recommendation scores}. This is where similar papers get seemingly random rankings from different reviewers, because they disagree on whether e.g. originality outweighs weaker evaluation. Even having a clear policy does not help\footnote{AAAI 2013 aimed to select the ``exciting but imperfect'' papers, and provided the reviewers with instructions about how to compute the overall recommendation based on individual rubric scores. However, they often ignored the instructions.} \cite{NoothigattuShahEtAl_2020_Loss_Functions_Axioms_and_Peer_Review}. The obvious solution is that reviewers should only be asked for specific scores (originality, technical soundness etc.), which would be the basis of the decisions according to the editorial policy.
    
The above solutions focus on reducing apples-to-oranges comparisons. A fundamentally different approach is to increase reviewer accountability, e.g. by making reviews public. Unfortunately, this does not address the core problem (reasoning under high uncertainty), and would introduce other problems.\footnote{Fundamentally, public reviews would force reviewers to spend more time to write more careful reviews. This would be great, but unless it is accompanied by systemic changes in how peer review is rewarded (which would not be quick or easy), it is likely to simply reduce participation. Public negative reviews also have repercussions for junior researchers.}

\section{What Holds Us Back?}

At this point, the reader might join the disappointed anonymous reviewers of this paper and say that we are not proposing anything new. This is precisely why the problem is so difficult: we lack the implementation, not the conceptual innovation -- and as researchers, we tend to only value the latter. 

On the organization side, each *ACL conference is organized by a new set of people each year who set their own policies. Such diversity by itself would be fantastic, but often, many things are changed at the same time, no systematic comparisons are drawn, and even the obviously successful innovations might not stay on. Consequently, next year we are no wiser about what works and what does not. We are running continuous experiments on ourselves, and never check\footnote{E.g. ACL 2020 opted to handle the increased reviewer load by making all authors register as reviewers, and EMNLP 2020 required a senior reviewer who would mentor secondary reviewers. How can we tell which strategy worked better and should be used next year?} the results.

\begin{figure}[!t]
    \centering
    \includegraphics[width=.65\columnwidth]{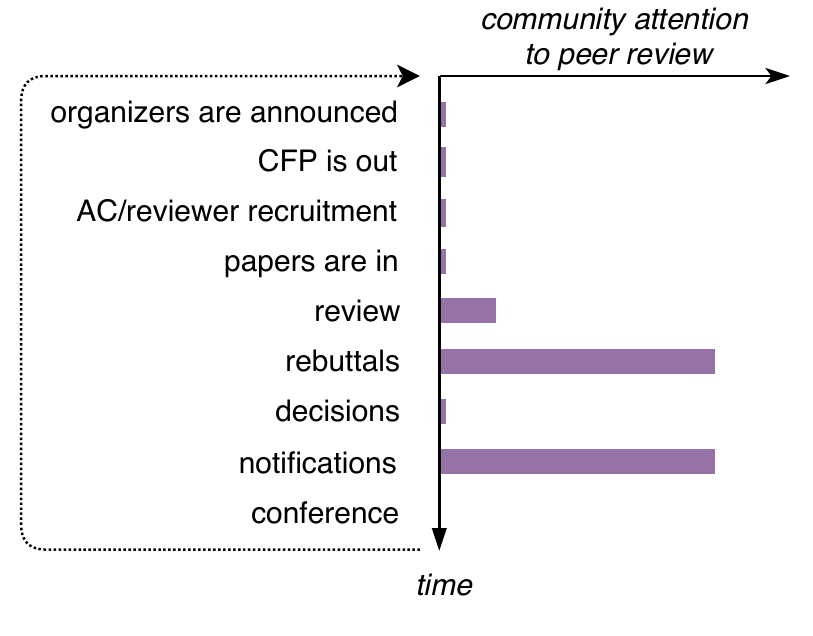}
    \caption{Attention to peer review in NLP community}
    \label{fig:process}
\end{figure}

On the community side, we are not aware of any quantitative studies of how peer review is discussed on social media, but as active members of the Twitter \#NLProc community, our impression is that this topic mostly gets on the feed during author rebuttals and after acceptance notifications at major conferences, as sketched in \autoref{fig:process}. At that time, there are bitter complaints and reform suggestions, but few practical initiatives (which ensures that the cycle is repeated at the next conference).

Fundamentally, peer review is an annotation problem, and we can try to tackle it because we know enough about experimental methodology, iterative guideline development, inter-annotator agreement, and biases. Here is where we fail:

\begin{itemize}[noitemsep,topsep=0pt]
    \item \textit{Organizers:} lack of mechanisms to test if one policy is better than another, and to ensure that successful policies are kept.
    \item \textit{Authors:} lack of willingness to actively monitor policy changes\footnote{For instance, \textit{Findings} was announced on the conference website and social media, but after acceptance notifications there was still confusion about what it meant.}, lack of ability to request reports on them and access the review data to conduct independent analysis.\footnote{Compulsory data collection opt-in for authors and reviewers is a less radical change than making all reviews public, and it might also reduce the number of one-line reviews.}
    \item \textit{Reviewers:} lack of recognition for meta-research as a valid part of NLP, which, as we learned in writing this paper, makes it difficult to publish on it. In a way, NLP peer review... prevents research on NLP peer review.
\end{itemize}

To illustrate the latter point: a quick search in the ACL anthology revealed only four conference papers on peer review from a meta-research perspective: a paper-reviewer matching tool \cite{AnjumGongEtAl_2019_PaRe_Paper-Reviewer_Matching_Approach_Using_Common_Topic_Space}, a corpus of reviews \cite{KangAmmarEtAl_2018_Dataset_of_Peer_Reviews_PeerRead_Collection_Insights_and_NLP_Applications}, and two experimental studies using NLP to explain the observed reviews \cite{CarageaUbanEtAl_2019_Myth_of_Double-Blind_Review_Revisited_ACL_vs_EMNLP,GaoEgerEtAl_2019_Does_My_Rebuttal_Matter_Insights_from_Major_NLP_Conference}. We could not find any ACL-published papers on testing different peer review policies or review form design for the NLP community. Yet without such work nothing will change, and no other field will do it for us.

The work by \citet{GaoEgerEtAl_2019_Does_My_Rebuttal_Matter_Insights_from_Major_NLP_Conference} offers an actionable insight: ACL reviewers appear to be victims of conformity bias, converging to the mean of reviews. One solution would be to let reviewers interact only with the authors during the rebuttal, but not with each other. The paper was published in NAACL -- yet, to the best of our knowledge, there have been no attempts to change any policies accordingly.

\section{Conclusion}

As a community familiar with annotation, we know that asking people to perform ill-defined tasks is not going to work well. Yet this is exactly what we expect of ourselves as reviewers. We can do better. 

There are many known ways to reduce uncertainty in paper merit estimation, such as improving the review forms and reviewer matching. The problem is that \textit{implementing} any of it would take a lot of work beside what ACL is already doing, sometimes counter to its current practices. Big changes in any large organization are difficult (especially in a volunteer-driven one), but this is the only way. 

The first step towards turning all the frustration on social media into action is to (a) recognize such work as respectable, main-track-worthy meta-research (so that there are incentives to do it at all), and (b) create new, voted-in ACL roles for systematic development, testing and comparison of review policies, as well as community feedback loops for authors and reviewers. A special ACL committee is working on a rolling review reform\footnote{\url{https://www.aclweb.org/adminwiki/index.php?title=ACL_Rolling_Review_Proposal}} to address the increasing \textit{volume} of reviews, but improving their \textit{quality} is a different, long-term project.

\section*{Acknowledgments}

This paper builds on numerous conversations, blog and Twitter posts. We are deeply indebted to Emily M.~Bender, Trevor Cohn, Leon Derczynski, Matt Gardner, Yoav Goldberg, Yang Liu, Tal Linzen, Ani Nenkova, Graham Neubig, Ted Pedersen, Ehud Reiter, Joseph M.~Rogers, Anna Rumshisky, Amanda Stent, Nathan Schneider, Karin Vespoor, Bonnie Webber, and many others. Many of the discussion points previously appeared on the Hacking Semantics blog.\footnote{https://hackingsemantics.xyz/}

We would also like to thank the anonymous reviewers for helping us to formulate the problem, and EMNLP organizers for excluding acknowledgements from the page limit.

\bibliography{emnlp2020}
\bibliographystyle{acl_natbib}

\end{document}